%% file: lrec2026-example.tex
\documentclass[10pt, a4paper]{article}

\usepackage[final]{lrec2026} % this is the new style
\usepackage{amsmath}
\usepackage{tikz}
\usepackage{subfig}
\usepackage{placeins}
\usetikzlibrary{positioning, arrows.meta, shapes.geometric}
\usepackage{afterpage}
\title{Scaling LLM Reasoning from Minimal Labels: A Semi-Supervised Framework with a Lightweight Verifier%Judging Reasoning to Build Reasoning: A Semi-Supervised Framework for Scaling LLM Reasoning from Minimal Supervision
}

\name{Keizo Kato\textsuperscript{1,2}, Chenhui Chu\textsuperscript{2}, Yugo Murawaki\textsuperscript{2}, Sado Kurohashi\textsuperscript{2,3}} 

\address{\textsuperscript{1}Fujitsu Limited, Kawasaki, Japan \\
\textsuperscript{2}Kyoto University, Kyoto, Japan \\
\textsuperscript{3}National Institute of Informatics, Tokyo, Japan \\
         }
\abstract{
For the development of Large language models (LLMs), recent approaches to generating pseudo intermediate reasoning have shown remarkable progress.
But they typically rely on large numbers of correctly annotated answers to assess reasoning quality. This paper presents a semi-supervised framework that scales reasoning learning from minimal supervision, turning reasoning verification itself into a data creation mechanism. We train a lightweight reasoning-correctness classifier on only a few labeled samples, which judges whether intermediate reasoning traces generated by an LLM are valid.
Furthermore, an entropy-based confidence threshold filters out unreliable samples, and the remaining high-confidence reasoning traces are used to fine-tune the model. Experiments on Verifiable Math Problems (Orca-Math subset) and Question Answering on Image Scene Graphs (GQA) with Visual Programming show that our method achieves accuracy comparable to using 10–15× more labeled data. Ablation analyses confirm that both the classifier and entropy filtering are essential for scalable and noise-resistant pseudo-labeling. By replacing expensive answer-level supervision with lightweight reasoning verification, our method provides a practical path toward constructing large-scale reasoning resources and paves the way for future autonomous reasoning systems that learn from minimal human input.
 \\ \newline  }

\begin{document}

\maketitleabstract
\section{Introduction}

Recent advances in Large Language Models (LLMs) have been driven not only by scaling laws but also by the explicit modeling of \textit{intermediate reasoning} processes. Such reasoning traces, often referred to as chains of thought, allow models to decompose complex problems into interpretable substeps, thereby improving both accuracy and robustness across various tasks. A series of studies---including Chain-of-Thought prompting~\citep{wei2022chain}, Self-Consistency~\citep{wang2023selfconsistency}, Tree-of-Thoughts~\citep{yao2023tree}, and ReAct~\citep{yao2023react}---have demonstrated that eliciting intermediate reasoning from LLMs significantly enhances performance on tasks requiring multi-step inference and logical composition.

However, constructing large-scale datasets containing intermediate reasoning annotations is prohibitively expensive and time-consuming. To address this challenge, a growing body of work has explored \textit{pseudo reasoning generation}, in which LLMs themselves generate intermediate reasoning traces that are then used for self-training. Methods such as Self-Training~\citep{zelikman2022selftraining}, Reflexion~\citep{shinn2023reflexion}, and Meta-CoT~\citep{xiang2025metacot} adopt the assumption that if an LLM’s final answer is correct, then its reasoning process leading to that answer can also be regarded as correct. This paradigm enables the automatic creation of reasoning-augmented data from question–answer pairs, alleviating the need for human-written explanations.

Nevertheless, these approaches still rely on the availability of \textit{answer annotations}. In many real-world domains---such as education, law, or medicine---obtaining correct answers requires expert verification and incurs substantial annotation costs. By contrast, \textit{problem statements themselves} are often abundant and inexpensive to collect, for example from textbooks, public Question and Answering (QA) corpora, or web-based question repositories. Thus, while questions are plentiful, answer labels remain a scarce and costly resource.

In this paper, we propose a novel training framework that enhances LLM performance even when only a few answer-annotated samples are available, leveraging a large pool of unlabeled question data. Specifically, we first train a \textbf{binary classifier} that determines whether an LLM-generated reasoning trace and its corresponding answer are correct, using only a small set of labeled examples. This classifier is then applied to unlabeled questions: the LLM generates intermediate reasoning and candidate answers, which are filtered by the classifier, and only those judged as ``correct'' are adopted as pseudo-labeled data. Furthermore, by introducing an \textbf{entropy-based confidence threshold}, we dynamically select reliable samples, leading to substantial performance gains with minimal supervision.

Our main contributions are threefold:
\begin{enumerate}
    \item We propose a learning framework that requires only a small number of answer-annotated samples while exploiting large-scale unlabeled question data to expand reasoning supervision.
    \item We introduce a classifier that judges the correctness of LLM-generated reasoning traces, enabling high-quality pseudo reasoning data selection.
    \item We demonstrate that entropy-based filtering substantially improves both accuracy and data efficiency, outperforming existing self-training methods in low-supervision scenarios.
\end{enumerate}
\input{fig_frame}
\input{related_work}
\input{method_new}
\input{experiment}
\input{conclusion}

\input{ethics}
\input{limiation}

\section{Bibliographical References}
\bibliographystyle{lrec2026-natbib}
\bibliography{lrec2026-example}

%\section{Language Resource References}
%\label{lr:ref}
\bibliographystylelanguageresource{lrec2026-natbib}
\bibliographylanguageresource{languageresource}

\end{document}

%% file: fig_frame.tex
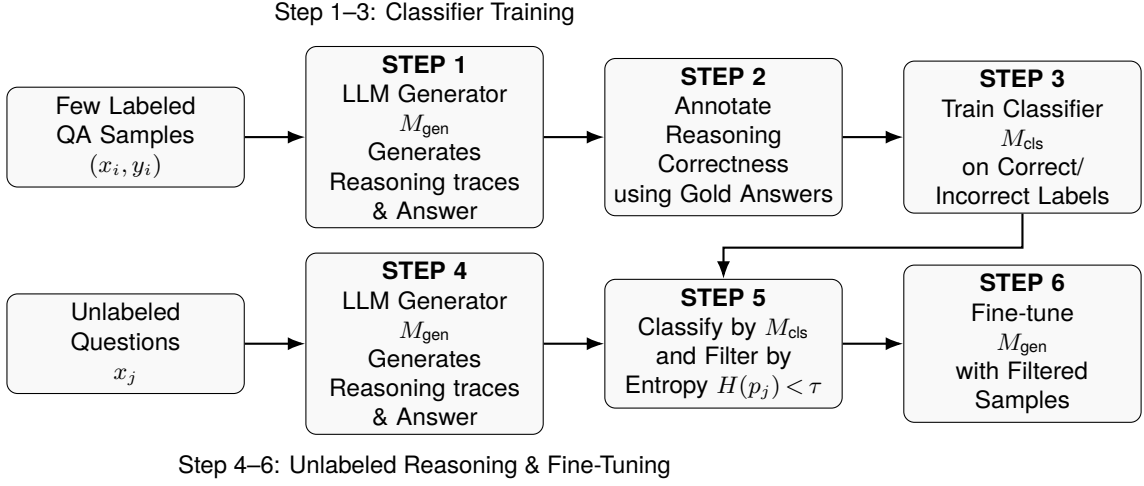
\begin{figure*}[t]
\centering
\begin{tikzpicture}[
    node distance=1.0cm and 0.8cm,
    every node/.style={font=\small},
    process/.style={
        rectangle, rounded corners, draw, align=center,
        minimum height=0.95cm, fill=gray!5,
        text width=2.9cm % ← 細身に
    },
    arrow/.style={-Latex, thick}
]

% --- Upper row: Step 1–3 (compact horizontal) -----------------
\node[process] (labeled) {Few Labeled\\QA Samples\\$(x_i, y_i)$};
\node[process, right=0.8cm of labeled] (gen_label) {\textbf{STEP 1} \\ LLM Generator\\$M_{\text{gen}}$\\Generates\\Reasoning traces \\ \& Answer};
\node[process, right=0.8cm of gen_label] (annotate) {\textbf{STEP 2} \\Annotate\\Reasoning Correctness\\using Gold Answers};
\node[process, right=0.8cm of annotate] (cls_train) {\textbf{STEP 3} \\Train Classifier\\$M_{\text{cls}}$\\on Correct/\\Incorrect Labels};

% --- Lower row: Step 4–6 (aligned under upper) ----------------
\node[process, below=1.4cm of labeled] (unlabeled) {Unlabeled\\Questions\\$x_j$};
\node[process, right=0.8cm of unlabeled] (gen_unlabeled) {\textbf{STEP 4}\\LLM Generator\\$M_{\text{gen}}$\\Generates\\Reasoning traces \\ \& Answer};
\node[process, right=0.8cm of gen_unlabeled] (cls_filter) {\textbf{STEP 5}\\Classify by $M_{\text{cls}}$\\and Filter by\\Entropy $H(p_j)\!<\!\tau$};
\node[process, right=0.8cm of cls_filter] (finetune) {\textbf{STEP 6}\\Fine-tune\\$M_{\text{gen}}$\\with Filtered\\Samples};

% --- Arrows: upper row (→) -----------------------------------
\draw[arrow] (labeled) -- (gen_label);
\draw[arrow] (gen_label) -- (annotate);
\draw[arrow] (annotate) -- (cls_train);

% --- Bridge down (short) -------------------------------------
\draw[arrow] (cls_train.south) -- ++(0,-0.45) -| (cls_filter.north);

% --- Lower row arrows (→) ------------------------------------
\draw[arrow] (unlabeled) -- (gen_unlabeled);
\draw[arrow] (gen_unlabeled) -- (cls_filter);
\draw[arrow] (cls_filter) -- (finetune);

% --- Step labels (compact) -----------------------------------
\node[above=0.15cm of gen_label, align=center]
    {Step 1–3: Classifier Training};
\node[below=0.15cm of gen_unlabeled,  align=center]
    {Step 4–6: Unlabeled Reasoning \& Fine-Tuning};
\end{tikzpicture}

\caption{
Overview of the proposed semi-supervised reasoning framework.
\textbf{Top (Step 1–3):} From a few labeled QA samples, $M_{\text{gen}}$ produces reasoning and answers;
reasoning correctness is annotated using gold answers, and a reasoning-correctness classifier $M_{\text{cls}}$ is trained.
\textbf{Bottom (Step 4–6):} For unlabeled questions, $M_{\text{gen}}$ generates reasoning; $M_{\text{cls}}$ classifies them and low-entropy (high-confidence) samples are retained to fine-tune $M_{\text{gen}}$.
}
\label{fig:framework}
\end{figure*}

%% file: related_work.tex
\section{Related Work}

%Our study is closely related to three major research areas:
%(1) reasoning enhancement via intermediate thoughts,
%(2) pseudo-label generation and self-training,
%(3) confidence-based filtering and sample selection under limited supervision,and (4) integration with self-verification and self-improvement frameworks.
Our study relates to reasoning enhancement with intermediate thoughts, pseudo-labeling for self-training, confidence-based sample selection, and self-improvement frameworks.
\subsection{Reasoning Enhancement via Intermediate Thoughts}
A growing body of research has shown that encouraging LLMs to produce explicit reasoning steps
improves their ability to solve complex problems.
Chain-of-Thought (CoT) prompting~\citep{wei2022chain} demonstrated that simple prompt engineering
can elicit step-by-step reasoning.
Self-Consistency~\citep{wang2023selfconsistency} enhanced reasoning stability
by sampling multiple reasoning traces and aggregating them through majority voting.
Tree-of-Thoughts~\citep{yao2023tree} proposed a search-based reasoning framework
that evaluates alternative reasoning branches systematically,
while ReAct~\citep{yao2023react} and Reflexion~\citep{shinn2023reflexion}
integrate reasoning, action, and self-correction for improved coherence.
Our work follows this line of research
but focuses on treating reasoning traces as data resources
whose quality can be automatically filtered and reused.

\subsection{Pseudo-Label Generation and Self-Training}
Since manually annotating reasoning steps is costly,
several studies have explored generating and reusing pseudo reasoning via self-training.
Self-Training for reasoning~\citep{zelikman2022selftraining} proposed iterative fine-tuning
on model-generated explanations,
while Meta-CoT~\citep{xiang2025metacot} introduced a meta-learning framework
to recursively improve reasoning generation.
Most of these methods assume that ``if the final answer is correct, the reasoning is also correct.''
Our approach relaxes this assumption by introducing a learned classifier
that discriminates between correct and incorrect reasoning,
thereby improving the robustness of pseudo-label selection.
In addition to correctness-based pseudo-labeling,
recent self-training approaches also exploit model confidence
to select reliable self-generated samples for further fine-tuning.
For instance, STaR~\citep{zelikman2022selftraining} and Meta-CoT~\citep{xiang2025metacot}
bootstrap reasoning ability from a few labeled examples
and reuse confident generations to improve performance iteratively.
However, these methods rely on the generator’s own confidence
(often measured by softmax probabilities or consistency across samples),
which can be unstable and prone to overconfidence in large language models.
In contrast, our framework decouples confidence estimation from generation
by introducing a separately trained reasoning-correctness classifier.
This external verifier provides a more objective confidence measure,
enabling more stable and transferable pseudo-label selection under minimal supervision.

\subsection{Confidence Estimation and Sample Selection}
Pseudo-labeling inherently risks introducing noisy or incorrect samples,
which can degrade performance.
To mitigate this, previous work has employed uncertainty-based filtering.
Confidence-based pseudo-labeling~\citep{arazo2020pseudo}
and uncertainty estimation frameworks~\citep{kendall2017uncertainties,alturki2025entropyfiltering}
often utilize entropy as a measure of reliability. 
Especially, \citet{saporta2020esl} introduced entropy-guided self-supervised learning for semantic segmentation, using entropy as a confidence measure for pseudo-label selection. They demonstrate that entropy-based filtering produces more reliable pseudo labels than maximum-probability thresholding.
Inspired by these studies, our framework applies an entropy-based confidence threshold
to the classifier’s predictions,
enabling the model to dynamically select high-confidence reasoning samples
while discarding unreliable ones.

\subsection{Integration with Self-Verification Frameworks}
Recent advances in self-verification and self-improvement of LLMs
have introduced iterative reasoning refinement loops.
Methods such as Self-Refine~\citep{madaan2023selfrefine},
ReGenesis~\citep{peng2025regenesis},
S$^2$R~\citep{ma2025s2r},
and Confident Reasoning~\citep{jang2025confidentreasoning}
allow LLMs to generate, critique, and revise their own reasoning.
However, most of these approaches let the same model act as both generator and verifier
(\textit{LLM-as-a-judge}) or rely on implicit confidence signals.
In contrast, our work explicitly trains a separate reasoning-correctness classifier,
which can be seamlessly incorporated into these frameworks
as a transferable, lightweight discriminator.

This design is not in conflict with prior methods; rather, it complements them.
By inserting an explicit reasoning-verification module into self-improvement loops,
one can stabilize pseudo-label generation,
suppress noise propagation,
and enable semi-supervised reasoning learning
even when explicit answer annotations are unavailable.
Therefore, our framework provides a modular component
that can further strengthen reinforcement-based reasoning optimization
(e.g., using GRPO or PPO)
and collaborative verification~\citep{liang2024improvingllmreasoningscaling},
bridging static pseudo-labeling and dynamic self-reinforcement.

%% file: method_new.tex
% =========================
% Section 3: Method (one-column friendly)
% =========================
\section{Method}
\label{sec:method}

\subsection{Overview and Intuition}
We aim to improve LLM reasoning from minimal supervision by \emph{learning to judge reasoning} and then \emph{using that judgment to scale training data}.  
The core intuition is twofold: (i) for LLMs, \textbf{binary verification} (``is this reasoning correct?'') is typically easier than \textbf{free-form generation} of correct reasoning; thus a small labeled set can train a useful verifier,\footnote{From cognitive psychology, \emph{recognition/verification} is generally simpler and more reliable than \emph{free recall/production} \citep{tenenbaum2011human}; our use of a binary verifier follows this asymmetry.} and (ii) \textbf{distributional uncertainty} (entropy) provides a more robust confidence signal than raw token probabilities, which are often overconfident and poorly calibrated~\citep{guo2017calibration,kendall2017uncertainties,desai2020calibration,kadavath2022lmknow}. 

The overview is shown in Figure \ref{fig:framework}. Operationally, we (1) generate reasoning on a few labeled QA pairs, (2) derive correctness labels from gold answers, (3) train an LLM-based correctness classifier, (4) generate reasoning on unlabeled problems, (5) filter by the classifier’s entropy, and (6) fine-tune the generator on the selected pseudo labels. 

An intuitive analogy is adversarial learning: in practice, a \emph{discriminator} (binary judge) often learns faster and provides a useful training signal to a harder \emph{generator}.  
This intuition is also aligned with the easy-to-hard generalization paradigm of \citet{sun2024easytohard}, which shows that an evaluation model trained on relatively simple, human-supervised tasks can be leveraged to progressively improve performance on more difficult tasks.
We adopt the same spirit here—first learn a lightweight judge for reasoning, then use its confidence to curate pseudo labels—while deferring broader theoretical evidence to Section~\ref{sec:theory-support}.

\subsection{Problem Setup and Notation}
Let \(\mathcal{D}_{\text{lab}}=\{(x_i,y_i)\}_{i=1}^{N_\ell}\) be a small labeled set (problem, gold answer) and \(\mathcal{D}_{\text{unlabeled}}=\{x_j\}_{j=1}^{N_u}\) a large unlabeled set.  
We use a generator \(M_{\text{gen}}\) and a verifier (classifier) \(M_{\text{cls}}\).  
A \emph{reasoning trace} is denoted \(r\), and a generated answer \(\hat{y}\).

\subsection{Training Steps}
\label{sec:method-actual}
\paragraph{Step 1: Reasoning Generation on Labeled Data.}
For each labeled pair \((x_i,y_i)\in\mathcal{D}_{\text{lab}}\), the generator produces a reasoning trace and a candidate answer as in Eq.~\eqref{eq:3_gen_labeled}:
\begin{equation}
r_i,\ \hat{y}_i \;=\; M_{\text{gen}}(x_i).
\label{eq:3_gen_labeled}
\end{equation}
This converts scarce labeled QA pairs into \((x_i,r_i,\hat{y}_i)\) triples that expose intermediate reasoning behavior.

\paragraph{Step 2: Correctness Annotation from Gold Answers.}
We derive a binary correctness label and encode it as a vocabulary token
\(z_i \in \{\texttt{true}, \texttt{false}\} \subset \mathcal{V}\)
(\texttt{true} if \(\hat{y}_i\) matches the gold answer \(y_i\), and \texttt{false} otherwise),
forming the dataset in Eq.~\eqref{eq:3_d_reason}:
\begin{equation}
\mathcal{D}_{\text{reason}}
\;=\;
\big\{(x_i,\, r_i,\, \hat{y}_i,\, z_i)\big\}_{i=1}^{N_\ell}.
\label{eq:3_d_reason}
\end{equation}
Turning generation into binary feedback provides a simpler signal that can be learned from few examples.

\paragraph{Step 3: Training a Reasoning-Correctness Classifier.}
We train \(M_{\text{cls}}\) to assign a distribution over the vocabulary
\(\mathcal{V}\), where \(v \in \mathcal{V}\) denotes a vocabulary token,
in Eq.~\eqref{eq:3_prob}:
\begin{equation}
p_i(v)
\;=\;
M_{\text{cls}}\!\big(v \mid x_i, r_i, \hat{y}_i\big).
\label{eq:3_prob}
\end{equation}
The objective is the cross-entropy loss over the vocabulary as in
Eq.~\eqref{eq:3_loss}:
\begin{equation}
\mathcal{L}_{\mathrm{cls}}
= - \sum_{\substack{(x_i,\, r_i,\, \hat{y}_i,\, z_i) \\ \in\, \mathcal{D}_{\mathrm{reason}}}}
\log p_i(z_i),
\label{eq:3_loss}
\end{equation}
where \(z_i \in \{\texttt{true}, \texttt{false}\} \subset \mathcal{V}\)
is the gold correctness token defined in Eq.~\eqref{eq:3_d_reason}.
This discriminative training calibrates relative confidence more reliably than relying on raw generation scores~\citep{desai2020calibration,kadavath2022lmknow}.

\paragraph{Step 4: Reasoning Generation on Unlabeled Problems.}
For each \(x_j\in\mathcal{D}_{\text{unlabeled}}\), the generator produces outputs as in Eq.~\eqref{eq:3_gen_unlabeled}:
\begin{equation}
r_j,\ \hat{y}_j \;=\; M_{\text{gen}}(x_j).
\label{eq:3_gen_unlabeled}
\end{equation}
The triples \((x_j, r_j, \hat{y}_j)\) are passed to the verifier in Step~5.

\paragraph{Step 5: Confidence-Based Filtering via Entropy.}
The classifier assigns a distribution over the vocabulary $\mathcal{V}$ for each
generated candidate:
\begin{equation}
  p_j(v) = M_{\mathrm{cls}}(v \mid x_j, r_j, \hat{y}_j), \quad v \in \mathcal{V},
  \label{eq:3_prob_unlabeled}
\end{equation}
and predicts correctness as
$z_j = \arg\max_{v \in \{\texttt{true},\texttt{false}\}} p_j(v)$.
For samples predicted as correct (i.e., $z_j = \texttt{true}$), we estimate
confidence using the entropy of the verifier's output distribution.
We compute this entropy in Eq.~\eqref{eq:3_entropy}:
\begin{equation}
  h(p_j) \;=\; -\sum_{v \in \mathcal{V}} p_j(v) \log p_j(v).
  \label{eq:3_entropy}
\end{equation}
Finally, we select samples based on Eq.~\eqref{eq:3_pseudo}:
\begin{equation}
\mathcal{D}_{\text{pseudo}}
\;=\;
\big\{
(x_j, r_j, \hat{y}_j)
\;\big|\;
z_j = \texttt{true},
\;
h(p_j) < \tau
\big\}.
\label{eq:3_pseudo}
\end{equation}
Entropy captures the verifier's uncertainty and yields higher-precision
pseudo labels in practice (Figure~\ref{fig:entropy_precision}; Sec.~\ref{sec:exp-filtering}).

\paragraph{Step 6: Fine-Tuning the Generator with Selected Pseudo Labels.}
We fine-tune \(M_{\text{gen}}\) on \(\mathcal{D}_{\text{lab}}\cup\mathcal{D}_{\text{pseudo}}\) via standard supervised fine-tuning (SFT), treating selected reasoning traces as training signals.  
In summary, the method converts a few labeled QA pairs into a reliable reasoning verifier, uses entropy-calibrated judgments to harvest high-precision pseudo reasoning from large unlabeled pools, and finally fine-tunes the generator on the selected set—realizing scalable, semi-supervised improvement of reasoning with minimal supervision.

% ---------------------------------------------------------------
% Place this at the end of the Method section
\subsection{Additional Theoretical Support}
\label{sec:theory-support}
Our design is further supported by prior findings that emphasize the relative tractability of \emph{verification/selection} over \emph{generation}.  
Self-Consistency improves performance by \emph{evaluating and selecting} among multiple reasoning traces rather than requiring a single perfect derivation~\citep{wang2023selfconsistency}.  
Self-Refine–style frameworks operationalize iterative \emph{critique and revision}, making verification a simpler subproblem that guides generation~\citep{madaan2023selfrefine}.  
In parallel, calibration studies indicate that large models provide \emph{useful relative confidence signals} even when textual generations are imperfect~\citep{guo2017calibration,desai2020calibration,kadavath2022lmknow}.  
Together, these results support our approach of (i) training a binary verifier from few labels and (ii) using its entropy-based confidence to select high-precision pseudo reasoning for scalable semi-supervised learning.

%% file: experiment.tex
\section{Experiments}
\input{table1}
\subsection{Experimental Setting and Motivation}

Our goal is to empirically verify whether the proposed framework can improve reasoning performance
under extremely limited annotation conditions.
Although our experiments use relatively small models and datasets,
they are designed to rigorously test the \textit{principle} of the method rather than to compete on scale.
Each experiment compares:
(1) SFT with a small number of labeled samples,
(2) our proposed semi-supervised framework with reasoning verification and entropy filtering,
and (3) SFT with a larger number of labeled samples as an upper bound.

\subsection{Dataset 1: Verifiable Math Problems (Orca-Math Subset)}
\label{sec:exp1}
\paragraph{Task Overview.}
We use the \textit{verifiable-math-problems} dataset on HuggingFace,
specifically the \textit{Orca-Math} subset, which contains mathematical reasoning problems
paired with intermediate reasoning and final answers.
We evaluate models using answer accuracy.

%\paragraph{Rationale for Selection.}
We adopt \textit{Orca-Math} as a representative mathematical reasoning benchmark
to examine whether our semi-supervised framework can improve reasoning ability
on general, language-based problem solving tasks that are widely used for evaluating large language models.
%Unlike domain-specific or multimodal reasoning settings, this dataset focuses purely on symbolic and textual reasoning, making it suitable for isolating the effects of reasoning verification.

Furthermore, several popular mathematical datasets,
including \textit{GSM8K} and \textit{MATH},
have been observed to show little or no improvement from SFT,
and previous works have raised concerns about possible data leakage
from pre-training corpora.
In contrast, \textit{Orca-Math} exhibits consistent performance gains under SFT,
suggesting that the influence of data leakage is relatively minor.
This makes it an appropriate and reliable testbed
for evaluating the effectiveness of semi-supervised reasoning learning
under limited supervision.

It is also worth noting that publicly available mathematical reasoning datasets
that are both general-purpose and largely unaffected by pre-training leakage
remain scarce.
In this context, \textit{Orca-Math} offers a balanced choice
between experimental clarity and general applicability.

\paragraph{Setup.}
We randomly split 500 samples for testing and 500 for validation.
We set the number of labeled QA pairs to 200 to balance \emph{practical} annotation budgets in expert domains and \emph{statistical} adequacy for a binary verifier. In practice, specialist supervision is costly and typically available only in the low hundreds—e.g., biomedical QA benchmarks such as \textit{PubMedQA}~\citep{jin2019pubmedqa} and the \textit{BioASQ} challenge~\citep{tsatsaronis2015bioasq} rely on expert involvement and limited high-quality labels; similarly, legal NLP benchmarks aggregate expert-curated tasks under constrained supervision (e.g., \textit{LexGLUE}~\citep{chalkidis2022lexglue}). Alignment-style supervision for general LMs is also carefully budgeted~\citep{ouyang2022instructgpt}. Taken together, few-hundred–scale seeds are a realistic starting point when labels must be expert-provided or statistically robust.\footnote{For a Bernoulli accuracy estimate of a binary verifier, a normal-approximation suggests $n \approx z_{1-\alpha/2}^2\,p(1-p)/\varepsilon^2$. Under the conservative worst case $p{=}0.5$ with $z_{0.975}{=}1.96$ and target margin $\varepsilon{=}0.07$, one obtains $n{\approx}196$ (about $200$), providing a stable validation/early-stopping signal without excessive annotation burden.}

The base model is Qwen2.5-0.5B-Instruct~\citep{qwen2.5}, a widely used open-source instruction-tuned LLM.  In this work, to investigate the self-improvement capabilities of LLMs, the same base model is employed for both the generator $M_{\text{gen}}$ and the classifier $M_{\text{cls}}$.
We intentionally adopt the smallest available configuration, as scenarios involving limited labeled data often coincide with low computational budgets in practical applications.  
Evaluating our method under such a constrained setting allows us to assess whether the proposed semi-supervised framework remains effective even when both annotation and computational resources are scarce.

We use the trl library~\citep{vonwerra2022trl} for training.  
For the reasoning-correctness classifier $M_{\text{cls}}$, early stopping is applied based on the classification accuracy on the validation set.  
For the final reasoning generator $M_{\text{gen}}$, early stopping is determined by the answer accuracy on the validation set.  
The batch size is set to $64$, the learning rate to $1\times10^{-6}$, and the warm-up ratio to $0.05$.  
These hyperparameters are empirically determined.
%This small-scale model allows us to focus on the learning dynamics without relying on massive capacity.

%\paragraph{Discussion.}
\paragraph{Results.}
The result is shown in Table~\ref{tab:math}. With only 200 labeled samples used to train the reasoning classifier,
the model filtered reasoning traces from 100,000 unlabeled problems
and selected the top 10\% most confident samples (based on entropy). This entropy threshold is determined by the observation described in Section \ref{sec:exp-filtering}.
This achieved 30.1\% accuracy—comparable to SFT using 3,000 labeled samples.
This result empirically supports our hypothesis that
LLMs can more easily learn to \textit{judge} reasoning correctness
than to generate reasoning itself.
Note that the same 200 labeled samples used for training the classifier
were also reused in the final model fine-tuning.

\input{table2}
\subsection{Dataset 2: GQA with Visual Programming}
\paragraph{Task Overview.}
To test the generality of our approach in a multi-modal setting,
we follow prior work on \textit{Visual Programming}~\citep{gupta2023visual,khan2024selftraining}.
GQA~\citep{hudson2019gqa} is a challenging Visual Question Answering (VQA) benchmark
requiring compositional reasoning over visual scenes.
This setup represents a case where the model must learn a new form of reasoning
that is not typically covered by general-purpose pre-training corpora,
thereby testing whether the proposed semi-supervised scheme
can extend LLM reasoning to previously unseen modalities and tasks.
Example of GQA with Visual Programming is shown as following.\\
\textbf{Question:} \emph{Are there both ties and glasses in the picture?}\\
\textbf{Program (pseudo-code):}
\begin{verbatim}
BOX0=Loc(image=IMAGE,object='tie')
ANSWER0=Count(box=BOX0)
BOX1=Loc(image=IMAGE,object='glass')
ANSWER1=Count(box=BOX1)
ANSWER2=Eval("yes" if (ANSWER0) > 0 
and (ANSWER1) > 0 else "no")
RESULT=ANSWER2
Prediction: no
\end{verbatim}

This example illustrates the visual program’s compositional structure:
object localization (\texttt{Loc}), quantitative queries (\texttt{Count}),
and symbolic combination via logical evaluation (\texttt{Eval}).
Our setting trains an LLM to generate such programs,
while a separate vision-language executor runs them to obtain the final answer.

Because the dataset does not include intermediate reasoning annotations,
direct SFT with reasoning supervision is not feasible.
Instead, we build on the self-training strategy of \citet{khan2024selftraining},
where pseudo-programs (object detection → attribute extraction → relation reasoning → answer)
are generated in a zero-shot manner, executed,
and only those that yield correct final answers are retained for fine-tuning.
Importantly, the final GQA question-answering task is solved by a
vision-language model separate from the program-generating LLM,
allowing us to isolate the quality of generated reasoning programs
from the perception and answering capabilities of the underlying visual model.

\paragraph{Setup.}
We adopt the same data split as prior work~\citep{khan2024selftraining}, using 1,912 samples for test
and 891 for validation. The number of labeled QA pairs is set to 200 as well in Section \ref{sec:exp1}.

The base model is Llama3-8B-Instruct~\citep{grattafiori2024llama3herdmodels} for both the generator $M_{\text{gen}}$ and the classifier $M_{\text{cls}}$.  Similar to Qwen2.5-0.5B used in the previous experiment, this model is a widely adopted open-source instruction-tuned LLM.
We select it because its scale is representative of commonly used mid-sized models, and evaluating our method on such a model allows us to assess the generality of the proposed framework across different architectures and parameter scales.

Zero-shot prompting uses 27 in-context examples as in \citet{khan2024selftraining}.
For our framework, we trained a reasoning-classification model
on 200 labeled samples and applied it to 91,300 unlabeled questions,
selecting the top 10\% most confident samples by entropy. This entropy threshold is determined by the observation described in Section \ref{sec:exp-filtering} as well.
%replaced to correct values later

We use the Hugging Face Trainer for training.  
For the reasoning-correctness classifier $M_{\text{cls}}$, early stopping is applied based on the classification accuracy on the validation set.  
For the reasoning generator $M_{\text{gen}}$ in the GQA setting, the final task accuracy depends on a separate vision-language execution engine; therefore, early stopping is determined by the validation loss of the generator itself rather than downstream answer accuracy. The batch size is set to $1$, the learning rate to $1\times10^{-3}$.  Gradient accumulation steps is set to $4$.
These hyperparameters are empirically determined. The rest of the hyperparameters are the default values for each library.
For executing the VQA task, we employ BLIP~\citep{blip} as the engine, and OwlVIT~\citep{owlvit} is used for object detection.
\input{figure1}
\paragraph{Results.}
%\paragraph{Discussion.}
Table~\ref{tab:gqa} summarizes the results of the baseline and proposed methods.
Our method improves over both zero-shot reasoning and the self-training baseline
with only 200 labeled samples,
approaching the performance of SFT using 2,000 labeled samples.
This demonstrates that the proposed reasoning-verification framework
can effectively identify and exploit reliable intermediate reasoning traces
even when the dataset lacks explicit annotations.
Importantly, this shows that the classifier-based filtering approach
is complementary to visual reasoning pipelines such as Visual Programming,
highlighting the potential of our method to generalize beyond textual reasoning tasks.
\subsection{Overall Discussion}

Although the experimental scale is modest,
the results consistently demonstrate the principle behind our approach:
%\begin{itemize}
    %\item 
    (i) Even with a very small number of answer annotations, reasoning quality can be improved through classifier-based filtering.
    %\item 
    (ii) The method achieves comparable gains to increasing labeled data by an order of magnitude.
    %\item 
    (iii)Intermediate reasoning can be treated as a renewable data resource, supporting efficient semi-supervised learning.
%\end{itemize}
Rather than emphasizing absolute performance,
we view these results as a \textit{proof-of-concept} confirming
that learning to judge reasoning correctness
is a feasible and efficient path toward scalable reasoning improvement,
especially in domains where labeling is costly.

\subsection{Effect of the Entropy-Based Confidence Filter}
\label{sec:exp-filtering}
A key finding across both datasets is that
the entropy-based confidence filtering mechanism
substantially improves the quality of pseudo-labeled reasoning samples.
To visualize this effect,
we analyze how the precision of selected samples changes
as the confidence threshold becomes stricter.
Figure~\ref{fig:entropy_precision} shows validation precision
when filtering by either the final-token probability (blue)
or the entropy of the classifier output (orange),
as low-confidence samples are progressively excluded
(i.e., retaining only the top-$k$\% most confident ones).

As illustrated, precision remains nearly flat
when filtering by raw probability,
indicating that the classifier’s predicted probabilities
are not well calibrated and tend to overestimate confidence.
In contrast, entropy-based filtering yields a clear and consistent improvement:
precision sharply increases when only the top 10\% of lowest-entropy samples are retained.
This behavior is observed in both (a) Orca-Math with Qwen2.5-0.5B
and (b) GQA with Llama3-8B,
demonstrating that entropy provides a more stable confidence signal
for pseudo-label selection across architectures and modalities.
This observation is consistent with the previous work in a different task domain \citep{saporta2020esl}.

\paragraph{Discussion.}
This analysis explains why entropy-based filtering was adopted in our main experiments.
Unlike raw probability thresholds, entropy better reflects distributional uncertainty, leading to higher-precision pseudo reasoning samples and more stable semi-supervised training behavior. The precision curves also provide a principled basis for selecting the 10\% high-confidence cutoff used throughout the paper.

\input{ablation}

%% file: table1.tex
%\afterpage{
\begin{table*}[h!]
\centering
\small
\begin{tabular}{lrrrrrr}
\hline
\textbf{Setting} & \textbf{Labeled}& \textbf{Unlabeled}& \textbf{Filtered}&\textbf{Total} & \textbf{Accuracy} \\
\hline
Zero-shot & – & - & - & - & 25.2 \\
SFT (200 labels) & 200 &0 & 0 & 200 & 26.6 \\
Proposed (200 labels + 100k unlabeled) & 200 & 100,000 & 2,911 & 3,111 & \textbf{30.1} \\ \hline
SFT (3,000 labels) & 3,000 & 0 & 0 & 3,000 & 30.6 \\
\hline
\end{tabular}
\caption{Results on Verifiable Math Problems (Orca-Math subset). \textbf{Labeled}, \textbf{Unlabeled}, \textbf{Filtered}, and \textbf{Total} indicate the number of samples. \textbf{Filtered} represents the number of samples remaining after filtering the \textbf{Unlabeled} data using  $M_{\text{cls}}$. \textbf{Total} is the final sample number for training (\textbf{Labeled} + \textbf{Filtered}).}
\label{tab:math}
\end{table*}
%}

%% file: table2.tex
\begin{table*}[h]
\centering
\small
\begin{tabular}{lrrrrr}
\hline
\textbf{Setting} & \textbf{Labeled} &\textbf{Unlabeled}& \textbf{Filtered}& \textbf{Total} &  \textbf{Accuracy} \\
\hline
Zero-shot & – & - & - & - & 52.2 \\
Khan et al. (SFT, 200 labels) & 200 & 0& 0& 200 &51.8 \\
Proposed (200 labels + 91k unlabeled) & 200  & 91,300 & 3,551 & 3,751 &\textbf{54.1} \\ \hline
Khan et al. (SFT, 2,000 labels) & 2,000 & 0 & 0 & 2,000 & 54.5 \\
\hline
\end{tabular}
\caption{Results on GQA with Visual Programming.}
\label{tab:gqa}
\end{table*}

%% file: figure1.tex
\begin{figure*}[th]
    \centering
    % --- (a) Orca-Math / Qwen ---
    \subfloat[Orca-Math (Qwen2.5-0.5B-Instruct)]{
        \includegraphics[width=0.49\linewidth]{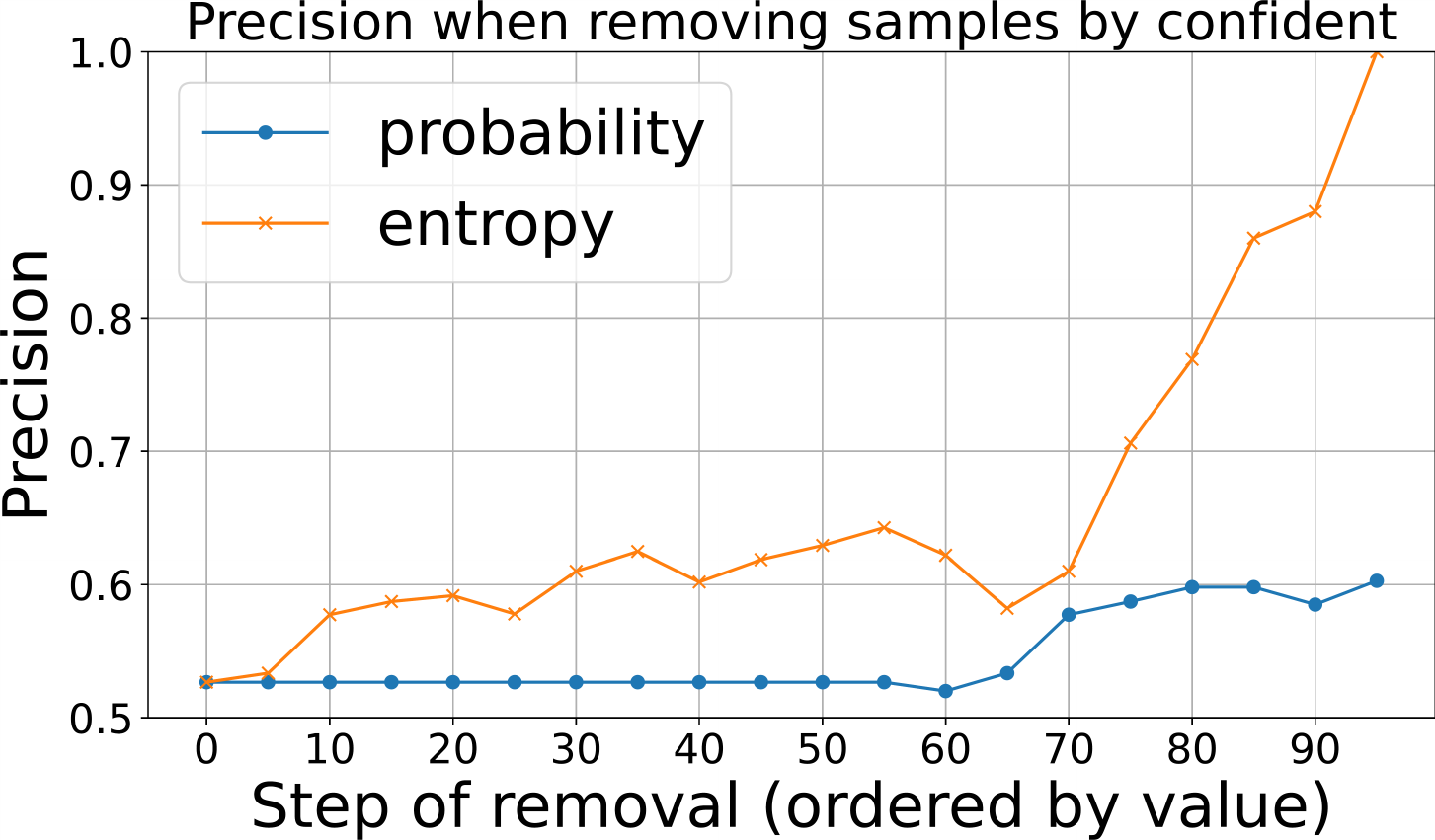}
    }%[-0.2em]
    % --- (b) GQA / Llama ---
    \subfloat[GQA (Llama3-8B-Instruct)]{
        \includegraphics[width=0.49\linewidth]{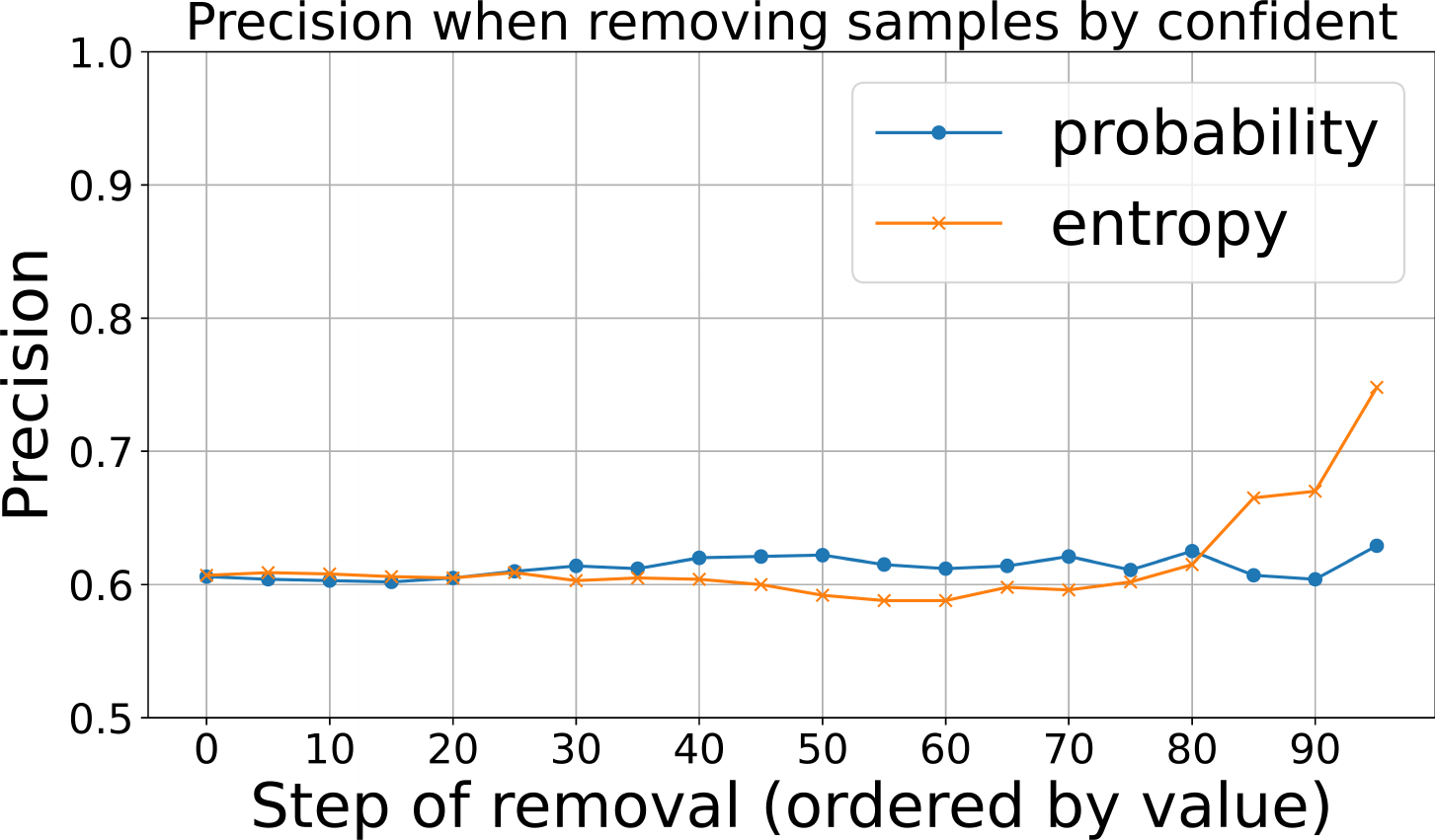}
    }
    \caption{
    Precision of pseudo-labeled reasoning samples under different confidence metrics.
    Each plot shows precision as a function of the exclusion ratio of low-confidence samples
    (x-axis: percentage of samples removed from least to most confident, y-axis: precision).
    Filtering by the final-token probability (blue) yields no or very small improvement,
    whereas entropy-based filtering (orange) produces a sharp precision rise
    around the 90\% exclusion point, where only the top 10\% of samples remain.
    }
    \label{fig:entropy_precision}
\end{figure*}

%% file: ablation.tex
\input{table3}
\input{table4}
\subsection{Ablation Study}
We further analyze the contribution of each component of our framework.
Specifically, we investigate:
(1) whether the reasoning-classifier itself is necessary,
(2) how the number of unlabeled samples affects performance, and
(3) the impact of the entropy-based confidence filter.

\subsubsection{Effect of the Reasoning Classifier and Entropy Filtering}

\paragraph{Verifiable Math Problems (Orca-Math).}
Table~\ref{tab:ablation_math} summarizes the results under different unlabeled data sizes and filtering conditions.
Without the reasoning classifier, simply augmenting training data with model-generated reasoning
does not lead to improvement.
Introducing the classifier already provides clear gains,
and adding entropy-based filtering further boosts accuracy, especially at larger unlabeled scales.

\paragraph{Discussion.}
The results clearly indicate that naive self-training without a reasoning-quality classifier (row 2)
fails to improve upon the zero-shot baseline.
In contrast, classifier-guided pseudo-labeling improves performance steadily with increasing unlabeled data.
Moreover, entropy-based filtering raises the performance ceiling:
it prevents degradation from noisy pseudo reasoning and enables higher precision in reasoning supervision.
Even when the overall number of samples is large,
without confidence filtering the improvement quickly saturates,
while the entropy filter continues to yield marginal gains.

\paragraph{GQA with Visual Programming.}
A similar trend is observed in the visual reasoning task (Table~\ref{tab:ablation_gqa}).
Because GQA does not contain explicit intermediate reasoning annotations,
the classifier acts as a crucial proxy for reasoning verification.
We again find that entropy filtering stabilizes training
and allows the model to selectively focus on reliable pseudo reasoning.

\paragraph{Discussion.}
For GQA, increasing unlabeled data improves performance only up to a certain saturation point.
Entropy filtering boosts the accuracy further by removing uncertain pseudo labels,
suggesting that the optimal trade-off between confidence threshold and data volume
depends on the task complexity and reasoning diversity.
Nevertheless, since question-only data are inexpensive to collect,
gradually expanding unlabeled samples with entropy-based filtering
is a practical way to scale reasoning learning in real-world settings.

\subsubsection{Summary of Findings}

Across both tasks, the following trends hold consistently:
%\begin{itemize}
    %\item 
    (1) Without the reasoning classifier, pseudo reasoning introduces noise and yields minimal benefit.
    %\item 
    (2) The reasoning classifier alone significantly improves data efficiency.
    %\item 
    (3) Entropy filtering raises the upper performance bound and mitigates saturation.
    %\item 
    (4) The trade-off between sample count and confidence threshold is task-dependent, but the general principle of selective pseudo-labeling remains robust.
%\end{itemize}

These ablations reinforce that our method’s strength lies not in brute-force data expansion,
but in leveraging LLMs’ inherent ability to \textit{discriminate} correct reasoning from incorrect reasoning,
enabling efficient semi-supervised scaling with minimal human annotation.

%% file: table3.tex
\begin{table*}[h]
\centering
\small
\begin{tabular}{lrrrrr}
\hline
\textbf{Setting} & \textbf{labeled} &\textbf{Unlabeled}& \textbf{Filtered}& \textbf{Total}& \textbf{Accuracy} \\
\hline
Zero-shot & – &-&-&-& 25.2 \\
w/o Classifier (100k) & 200 & 100,000& 100,000 & 100,200 & 25.6 \\
Classifier (10k) &200 & 10,000 & 3,440 & 3,640 & 27.2 \\
Classifier (10k, top 10\%) &200& 10,000 & 285 & 305 & 28.2 \\
Classifier (100k) & 200 & 100,000 & 34,731 & 34,931 & 27.4 \\
Classifier (100k, top 10\%) &200& 100,000 & 2,911 &3,111& \textbf{30.1} \\
\hline
\end{tabular}
\caption{
Ablation results on Verifiable Math Problems (Orca-Math subset).
The reasoning classifier and entropy-based filtering consistently improve accuracy.
}
\label{tab:ablation_math}
\end{table*}

%% file: table4.tex
\begin{table*}[h]
\centering
\small
\begin{tabular}{lrrrrr}
\hline
\textbf{Setting} & \textbf{labeled} &\textbf{Unlabeled}& \textbf{Filtered}& \textbf{Total}& \textbf{Accuracy} \\
\hline
Zero-shot & – & - & - & -& 52.2 \\
w/o Classifier (91k) &200& 91,130& 91,130& 91,330 & 52.0 \\
Classifier (9k) &200 & 9,113 &3,923 & 4,123& 53.4 \\
Classifier (9k, top 10\%) &200 & 9,113 & 338& 538 &52.2 \\
Classifier (91k) & 200 & 91,130 & 42,152& 42,352 & 53.6 \\
Classifier (91k, top 10\%) &200& 91,130 & 3,551&3,751&\textbf{54.1} \\
\hline
\end{tabular}
\caption{
Ablation results on GQA with Visual Programming.
Entropy filtering enhances data efficiency, particularly when scaling to large unlabeled corpora.
}
\label{tab:ablation_gqa}
\end{table*}

%% file: conclusion.tex
\section{Conclusion}
This paper presents a semi-supervised framework that improves LLM reasoning with minimal supervision by \emph{learning to judge reasoning}: a correctness classifier, trained on few labeled examples, filters reliable reasoning traces from large unlabeled pools. On textual (Orca-Math) and multimodal (GQA) tasks, the method matches the performance of using 10–15$\times$ more labeled data, indicating that \emph{judging} is a data-efficient proxy for \emph{generating} reasoning.

Our ablation studies further revealed that:
%\begin{itemize}
%\item 
(1) without the classifier, naive pseudo-labeling brings little gain;
%\item 
(2) entropy-based filtering prevents saturation and raises the upper performance bound; and
%\item 
(3) the optimal balance between confidence and quantity depends on task complexity.
%\end{itemize}
Together, these results highlight that selective self-training—guided by confidence and reasoning discrimination—
is more effective than brute-force data expansion.

\paragraph{Future Outlook.}
We envision autonomous reasoning learning systems capable of self-improvement through problem generation, reasoning, and verification, eliminating human annotation. This closed-loop approach, integrating reasoning verification with automatic problem generation, offers a scalable path for self-evolving LLMs to achieve lifelong reasoning acquisition.

%% file: ethics.tex
\section*{Ethics Statement}
We use the public datasets Verifiable Math Problems (Orca-Math subset) and GQA strictly under their licenses for research purposes. Both datasets are provided without personally identifiable information; problem statements, (when applicable) intermediate reasoning, and final answers are already included in the datasets, and we did not conduct any new human annotation. Because data leakage into pre-training corpora is a known concern for mathematical benchmarks, we primarily adopt Orca-Math, where SFT yields observable gains, to comparatively reduce leakage effects.
While our method can expand pseudo reasoning from few labels, it can also propagate erroneous or biased reasoning. We mitigate this risk via an explicit classifier and entropy-based confidence filtering. However, for applications in high-stakes domains (e.g., medical or legal), human oversight and validation are strongly recommended.

%% file: limiation.tex
\section*{Limitations}

While our study demonstrates the potential of reasoning verification
for semi-supervised learning with large language models,
several limitations should be acknowledged.

\paragraph{Experimental Scale.}
The experiments were conducted on relatively small models
(\textit{Qwen2.5-0.5B-Instruct} and \textit{Llama3-8B-Instruct})
and limited datasets (Orca-Math and GQA subsets).
Our goal was to validate the core principle rather than absolute performance.
Nevertheless, larger-scale evaluations are necessary to confirm
the generality and robustness of the proposed framework across domains and architectures.

\paragraph{Entropy-Based Thresholding.}
The confidence estimation component in this work
relies solely on Shannon entropy as a simple uncertainty measure.
Although effective, this criterion does not account for model calibration,
multi-modal uncertainty, or hierarchical dependencies among reasoning steps.
Future work could explore adaptive thresholding,
Bayesian confidence estimation, or energy-based filtering
to better balance coverage and precision in pseudo reasoning selection.

\paragraph{Learning Framework.}
The proposed method is implemented in a supervised fine-tuning (SFT) setting,
without incorporating reinforcement signals.
While this choice ensures stability and reproducibility,
the framework could be extended to reinforcement learning formulations
such as GRPO or PPO-based reasoning optimization,
where the reasoning classifier serves as a reward model.
Integrating reasoning verification into reinforcement loops
may allow continuous self-improvement beyond static pseudo labeling.

\paragraph{Scope of Reasoning.}
Finally, our approach has been tested only on tasks
where intermediate reasoning is expressed in textual or structured form.
Extending this paradigm to free-form or multi-modal reasoning
(e.g., diagrammatic or procedural reasoning)
requires further investigation into how classifiers interpret and evaluate
non-textual reasoning traces.

Despite these limitations,
we believe the simplicity and extensibility of the proposed framework
make it a promising foundation for future research
in scalable, self-supervised reasoning systems.